\documentclass{article}
\usepackage{spconf,amsmath,graphicx}
\usepackage{multicol}
\usepackage{multirow}
\usepackage{subfigure}
\newcommand{\minitab}[2][l]{\begin{tabular}{#1}#2\end{tabular}}

\title{Multi-scale convolutional neural networks for crowd counting}
\name{Lingke Zeng, Xiangmin Xu\sthanks{Xiangmin Xu is the corresponding author.}, Bolun Cai, Suo Qiu, Tong Zhang}
\address{School of Electronic and Information Engineering\\
South China University of Technology, Guangzhou, China\\
z.lineker@gmail.com, xmxu@scut.edu.cn, caibolun@gmail.com, q.suo@foxmail.com, tony@scut.edu.cn}
\begin{document}
%
\maketitle
\begin{abstract}
Crowd counting on static images is a challenging problem due to scale variations. Recently deep neural networks have been shown to be effective in this task. However, existing neural-networks-based methods often use the multi-column or multi-network model to extract the scale-relevant features, which is more complicated for optimization and computation wasting. To this end, we propose a novel multi-scale convolutional neural network (MSCNN) for single image crowd counting. Based on the multi-scale blobs, the network is able to generate scale-relevant features for higher crowd counting performances in a single-column architecture, which is both accuracy and cost effective for practical applications. Complemental results show that our method outperforms the state-of-the-art methods on both accuracy and robustness with far less number of parameters.

\end{abstract}


\section{Introduction}
\label{secsec:intro:introsec:intro}
Crowd counting aims to estimate the number of people in the crowded images or videos feed from surveillance cameras. Overcrowding in scenarios such as tourist attractions and public rallies can cause crowd crushes, blockages and even stampedes. It has been much significant to public safety to produce an accurate and robust crowd count estimation using computer vision techniques. 

Existing methods of crowd counting can be generally divided into two categories: \emph{detection-based} methods and \emph{regression-based} methods.

\emph{Detection-based} methods generally assume that each person on the crowd images can be detected and located by using the given visual object detector \cite{C1,C2,C3}, and obtain the counting result by accumulating each detected person. However, these methods \cite{C4, C5, C6} need huge computing resource and they are often limited by person occlusions and complex background in practical scenarios, resulting at a relatively low robustness and accuracy.

\emph{Regression-based} methods regress the crowd count from the image directly. Chan et al. \cite{C7} used handcraft features to translate the crowd counting task into a regression problem. Following works \cite{C8,C9} proposed more kinds of crowd-relevant features including segment-based features, structural-based features and local texture features. Lempitsky et al. \cite{C10} proposed a density-based algorithm that obtain the count by integrating the estimated density map.

Recently, deep convolutional neural networks have been shown to be effective in crowd counting. Zhang et al. \cite{C11} proposed a convolutional neural network (CNN) to alternatively learn the crowd density and the crowd count. Wang et al. \cite{C12} directly used a CNN-based model to map the image patch to its people count value. However, these single-CNN-based algorithms are limited to extract scale-relevant features and hard to address the scale variations on crowd images. Zhang et al. \cite{C13} proposed a multi-column CNN to extract multi-scale features by columns with different kernel sizes. Boominathan et al. \cite{C14} proposed a multi-network CNN that used a deep and shallow network to improve the spatial resolution. These improved algorithms can relatively suppress the scale variations problem, but they still have two shortages:
 \begin{itemize}
   \item Multi-column/network need pre-trained single-network for global optimization, which is more complicated than end-to-end training.
   \item Multi-column/network introduce more parameters to consume more computing resource, which make it hard for practical application.
 \end{itemize}

In this paper, we propose a \textbf{multi-scale convolutional neural network (MSCNN)} to extract scale-relevant features. Rather than adding more columns or networks, we only introduce a multi-scale blob with different kernel sizes similar to the naive Inception module \cite{C15}. Our approach outperforms the state-of-the-art methods on the ShanghaiTech and UCF\_CC\_50 dataset with a small number of parameters.

\section{MULTI-SCALE CNN FOR CROWD COUNTING}
\label{sec:multiscalcnn}

\begin{figure*}[htb]
	\centering
	\centerline{\includegraphics[width=1\linewidth]{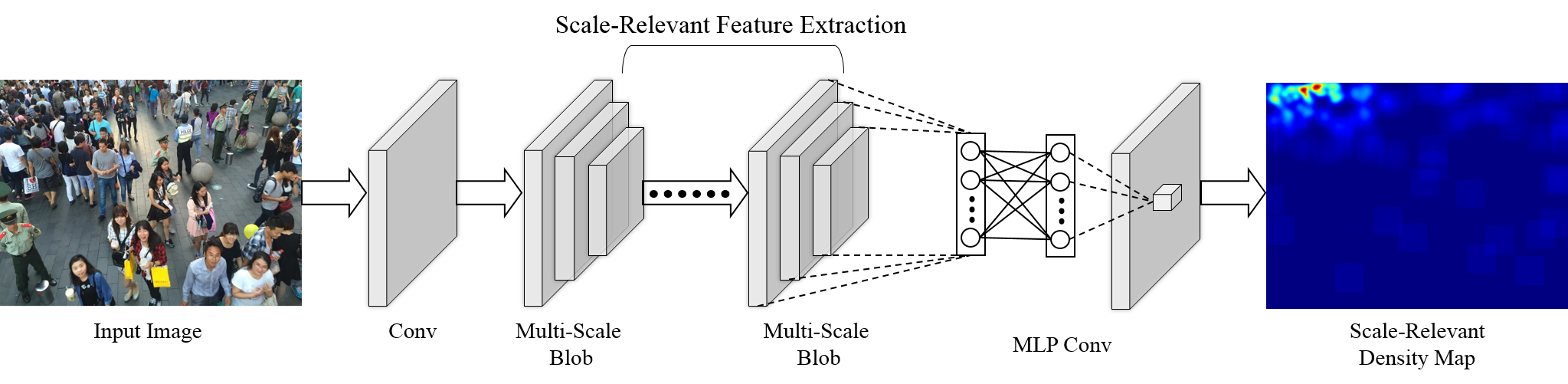}}
	\caption{Multi-scale convolutional neural network for crowd counting.}
	\label{fig:mscnn}
\end{figure*}

Crowd images are usually consisted of various sizes of person¡¯s pixels due to perspective distortion. Single-network is hard to counter scale variations with the same sized kernels combination. In \cite{C15}, a Inception module is proposed to process visual information at various scales and aggregated to the next stage. Motivated by it, we designed a multi-scale convolutional neural network (MSCNN) to learn the scale-relevant density maps from original images.

\subsection{Multi-scale Network Architecture}
\label{ssec:architecture}
\begin{figure}[h]
\centering
\centerline{\includegraphics[width=0.9\linewidth]{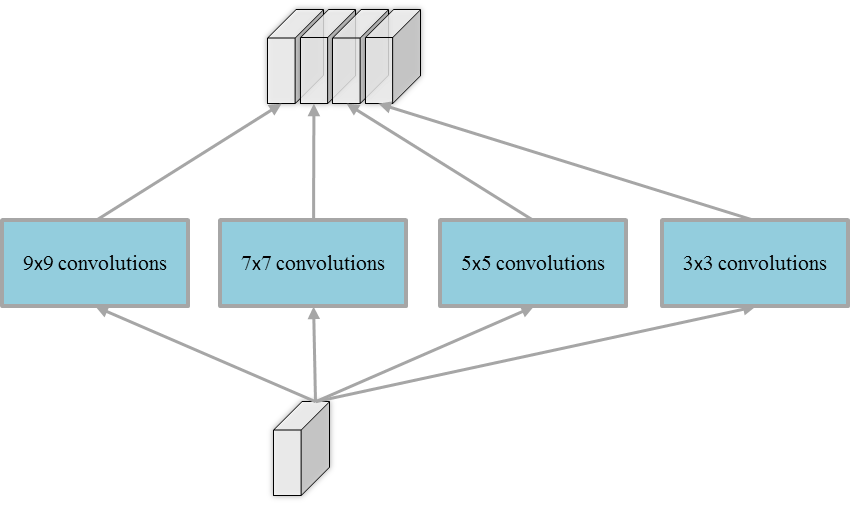}}
\caption{Multi-scale blob with different kernel size.}
\label{fig:icneption}
\end{figure}

An overview of MSCNN is illustrated in Figure. \ref{fig:mscnn}, including \emph{feature remapping}, \emph{multi-scale feature} extraction, and \emph{density map regression}. The first convolution layer is a traditional convolutional layer with single-sized kernels to remap the image feature. Multi-Scale Blob (MSB) is a Inception-like model (as Figure. \ref{fig:icneption}) to extract the scale-relevant features, which consists of multiple filters with different kernel size (including 9$\times$9, 7$\times$7, 5$\times$5 and 3$\times$3). A multi-layer perceptron (MLP) \cite{nin} convolution layer works as a pixel-wise fully connection, which has multiple $1\times1$ convolutional filters to regress the density map. Rectified linear unit (ReLU) \cite{C16} is applied after each convolution layer, which works as the activation function of previous convolutional layers except the last one. Since the value in density map is always positive, adding ReLU after last convolutional layer can enhance the density map restoration. Detailed parameter settings are listed in Table \ref{tab:modelarchitecture}.
\begin{table}[h]
  \centering
  \caption{The multi-scale CNN architecture.}\label{tab:modelarchitecture}
   \footnotesize
    \begin{tabular}{|c|c|c|c|c|}
    \hline
    {\bf Formulation} & {\bf Type} & {\bf Num.} & {\bf Filter Size} & {\bf Pad} \\
    \hline
    \multirow{2}{*}{\minitab[c]{Feature\\Remap}} & Conv & 64 & {9$\times$9} & 4 \\
    & ReLU & - & - & - \\ \hline
    \multirow{2}{*}{\minitab[c]{Multi-scale\\Feature}} & MSB Conv & 4$\times$16 & (9/7/5/3)$\times$(9/7/5/3) & 4/3/2/1 \\
     & ReLU & - & - & - \\ \hline
    Down-sample & MAX Pool & - & 2$\times$2 & 0 \\ \hline
    \multirow{4}{*}{\minitab[c]{Multi-scale\\Feature}} & MSB Conv & 4$\times$32 & (9/7/5/3)$\times$(9/7/5/3) & 4/3/2/1 \\
    & ReLU & - & - & - \\
    & MSB Conv & 4$\times$32 & (9/7/5/3)$\times$(9/7/5/3) & 4/3/2/1 \\
    & ReLU & - & - & - \\ \hline
    Down-sample & MAX Pool & - & 2$\times$2 & 0 \\ \hline
    \multirow{4}{*}{\minitab[c]{Multi-scale\\Feature}} &MSB Conv & 3$\times$64 & (7/5/3)$\times$(7/5/3) & 3/2/1 \\
    & ReLU & - & - & - \\
    &MSB Conv & 3$\times$64 & (7/5/3)$\times$(7/5/3) & 3/2/1 \\
    & ReLU & - & - & - \\ \hline
    \multirow{4}{*}{\minitab[c]{Density Map\\Regression}} & MLP Conv & 1000 & {1$\times$1} & 0 \\
    & ReLU & - & - & - \\
    & Conv & 1 & {1$\times$1} & 0  \\
    & ReLU & - & - & - \\ \hline
    \end{tabular}
\end{table}

\subsection{Scale-relevant Density Map}
\label{ssec:densitymap}

Following Zhang et al. \cite{C13}, we estimate the crowd density map directly from the input image. To generate a scale-relevant density map with high quality, the scale-adaptive kernel is currently the best choice. For each head annotation of the image, we represent it as a delta function $\delta\left(x - x_i\right)$ and describe its distribution with a Gaussian kernel $G_\sigma$ so that the density map can be represented as $F\left(x\right)=H\left(x\right)*G_\sigma\left(x\right)$ and finally accumulated to the crowd count value. If we assume that the crowd is evenly distributed on the ground plane, the average distance $\overline{d_i}$ between the head $x_i$ and its nearest 10 annotations can generally characterize the geometric distortion caused by perspective effect using the Eq. \eqref{eq:f}, where $M$ is the total number of head annotations in the image and we fix $\beta=0.3$ as \cite{C13} empirically.
\begin{equation}\label{eq:f}
F\left(x\right)=\sum_{i=1}^{M}\delta \left(x-x_i\right )*G_{\sigma_i}, \; \text{with} \; \sigma_i=\beta\overline{d_i}
\end{equation}
\begin{figure*}[htp]
\centering
\begin{minipage}[b]{0.33\linewidth}
  \centering
  \centerline{\includegraphics[width=4.5cm]{parta_img.jpg}}
  \small\centerline{(a) \emph{Part\_A} Test image}\medskip
\end{minipage}
\begin{minipage}[b]{0.33\linewidth}
  \centering
  \centerline{\includegraphics[width=4.5cm]{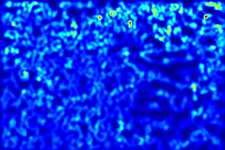}}
  \small\centerline{Ground truth: 1603}\medskip
\end{minipage}
\begin{minipage}[b]{0.33\linewidth}
  \centering
  \centerline{\includegraphics[width=4.5cm]{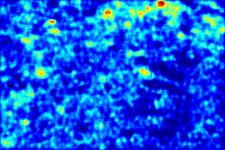}}
  \small\centerline{Estimation: 1687}\medskip
\end{minipage}

\begin{minipage}[b]{0.33\linewidth}
  \centering
  \centerline{\includegraphics[width=4.5cm]{partb_img.jpg}}
  \small\centerline{(b) \emph{Part\_B} Test image}\medskip
\end{minipage}
\begin{minipage}[b]{0.33\linewidth}
  \centering
  \centerline{\includegraphics[width=4.5cm]{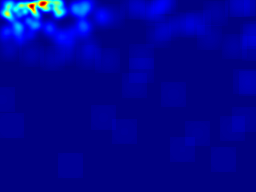}}
  \small\centerline{Ground truth: 104}\medskip
\end{minipage}
\begin{minipage}[b]{0.33\linewidth}
  \centering
  \centerline{\includegraphics[width=4.5cm]{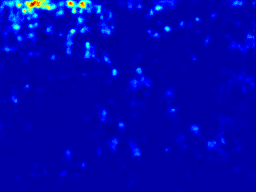}}
  \small\centerline{Estimation: 105}\medskip
\end{minipage}

\caption{The ground truth and estimated density map of test images in ShanghaiTech dataset}
\label{fig:res}
\end{figure*}
\subsection{Model Optimization}
\label{ssec:optimize}
The output from our model is mapped to the density map, Euclidean distance is used to measure the difference between the output feature map and the corresponding ground truth. The loss function that needs to be optimized is defined as Eq. \eqref{eq:L}, where $\Theta$ represents the parameters of the model while $F\left(X_i; \Theta\right)$ represents the output of the model. $X_i$ and $F_i$ are respectively the $i^{th}$ input image and density map ground truth.
\begin{equation}\label{eq:L}
L\left(\Theta\right)=\frac{1}{2N}\sum_{i=1}^{N}\left\|F\left(X_i; \Theta\right)-F_i \right\|_{2}^{2}
\end{equation}

\section{EXPERIMENTS}
\label{sec:experiemts}
We evaluate our multi-scale convolutional neural network (MSCNN) for crowd counting on two different datasets, which include the ShanghaiTech and UCF\_CC\_50 datasets. The experimental results show that our MSCNN outperforms the state-of-the-art methods on both accuracy and robustness with far less parameter. All of the convolutional neural networks are trained based on {\em{Caffe}} \cite{C17}.

\subsection{Evaluation Metric}
\label{ssec:evaluate}
Following existing state-of-the-art methods \cite{C13}, we use the mean absolute error (MAE), the mean squared error (MSE) and the number of neural network¡¯s parameters (PARAMS) to evaluate the performance on the testing datasets. The MAE and the MSE are defined in Eq. \eqref{eq:mae} and Eq. \eqref{eq:mse}.
\begin{equation}\label{eq:mae}
MAE=\frac{1}{N}\sum_{i=1}^{N}\left|z_i-\hat{z}_i\right|
\end{equation}
\begin{equation}\label{eq:mse}
MSE=\sqrt{\frac{1}{N}\sum_{i=1}^{N}\left(z_i-\hat{z}_i\right)^{2}}
\end{equation}
Here $N$ represents the total number of images in the testing datasets, $z_i$ and $\hat{z}_i$ are the ground truth and the estimated value respectively for the $i^{th}$ image. In general, MAE, MSE and PARAMS can respectively indicate the accuracy, robustness and computation complexity of a method.

\subsection{The ShanghaiTech Dataset}
\label{ssec:expshanghai}
The ShanghaiTech dataset is a large-scale crowd counting dataset introduced by \cite{C13}. It contains 1198 annotated images with a total of 330,165 persons. The dataset consists of 2 parts: \emph{Part\_A} has 482 images crawled from the Internet and \emph{Part\_B} has 716 images taken from the busy streets. Following \cite{C13}, both of them are divided into a training set with 300 images and a testing set with the remainder.

\subsubsection{Model Training }
\label{sssec:shanghaidataaug}
To ensure a sufficient number of data for model training, we perform data augmentation by cropping 9 patches from each image and flipping them. We simply fix the 9 cropped points as top, center and bottom combining with left, center and right. Each patch is 90\% of the original size.

In order to facilitate comparison with MCNN architecture \cite{C13}, the network was designed similar to the largest column of MCNN but with MSB, of which detailed settings are described in Table \ref{tab:modelarchitecture}. All convolutional kernels are initialized with Gaussian weight setting standard deviation to 0.01. As described in Sec. \ref{ssec:optimize}, we use the SGD optimization with momentum of 0.9 and weight decay as 0.0005.

\subsubsection{Results}
\label{sssec:shanghaiexp}
We compare our method with 4 existing methods on the ShanghaiTech dataset. The LBP+RR method used LBP feature to regress the function between the counting value and the input image. Zhang et al. \cite{C11} designed a convolutional network to regress both the density map and the crowd count value from original pixels. A multi-column CNN \cite{C13} is proposed to estimate the crowd count value (MCNN-CCR) and crowd density map (MCNN).

In Table \ref{tab:shanhai}, the results illustrate that our approach achieves the state-of-the-art performance on the ShanghaiTech dataset. In addition, it should be emphasized that the number of our parameters is far less than other two CNN-based algorithms. MSCNN uses approximately \textbf{7$\times$} fewer parameters than the state-of-the-art method (MCNN) with higher accuracy and robustness.
\begin{table}[h]
  \centering
  \caption{Performances of methods on ShanghaiTech dataset.}
  \label{tab:shanhai}
   \footnotesize
    \begin{tabular}{|c|c|c|c|c|c|}
    \hline
    \multirow{2}{*}{\bf Method} & \multicolumn{2}{c|}{\bf Part\_A} & \multicolumn{2}{c|}{\bf Part\_B} & \multirow{2}{*}{\bf PARAMS}\\
    \cline {2-5}
     &{\bf MAE} &{\bf MSE} &{\bf MAE} & {\bf MSE} & {}\\
    \hline
    {LBP+RR}&303.2&371.0&59.1&81.7&-\\
    \hline
    {MCNN-CCR \cite{C13}}&245.0&336.1&70.9&95.9&-\\
    \hline
    {Zhang et al. \cite{C11}}&181.8&277.7&32.0&49.8&7.1M\\
    \hline
    {MCNN \cite{C13}}&110.2&173.2&26.4&41.3&19.2M\\
    \hline
    {MSCNN}&{\bf 83.8}&{\bf 127.4}&{\bf 17.7}&{\bf 30.2}&{\bf 2.9M}\\
    \hline
    \end{tabular}
\end{table}
\subsection{The UCF\_CC\_50 Dataset}
\label{ssec:expucf}
The UCF\_CC\_50 dataset \cite{C18} contains 50 gray scale images with a total 63,974 annotated persons. The number of people range from 94 to 4543 with an average 1280 individuals per image. Following \cite{C11, C13, C14}, we divide the dataset into five splits evenly so that each split contains 10 images. Then we use 5-fold cross-validation to evaluate the performance of our proposed method.

\subsubsection{Model Training}
\label{sssec:ucfdtrain}
The most challenging problem of the UCF\_CC\_50 dataset is the limited number of images for training while the people count in the images span too large. To ensure enough number of training data, we perform a data augmentation strategy following \cite{C14} by randomly cropping 36 patches with size 225$\times$225 from each image and flipping them as similar in Sec. \ref{sssec:shanghaidataaug}.

We train 5 models using 5 splits of training set. The MAE and the MSE are calculated after all the 5 models obtained the estimated results of the corresponding validation set. During training, the MSCNN model is initialized almost the same as the experiment on the ShanghaiTech dataset except that the learning rate is fixed to be 1e-7 to guarantee the model convergence.

\subsubsection{Results}
\label{sssec:ucfexp}
We compared our method on the UCF\_CC\_50 dataset with 6 existing methods. In \cite{C19,C10,C18}, handcraft features are used to regress the density map from the input image. Three CNN-based methods \cite{C11,C14,C13} proposed to used multi-column/network and perform evaluation on the UCF\_CC\_50 dataset.

Table \ref{tab:ucf_comparison} illustrates that our approach also achieves the state-of-the-art performance on the UCF\_CC\_50 dataset. Here our parameters number is approximately \textbf{5$\times$} fewer than the CrowdNet model, demonstrating that our proposed MSCNN can work more accurately and robustly.
\begin{table}[h]
  \centering
  \caption{Performances of methods on UCF\_CC\_50 dataset.}
  \label{tab:ucf_comparison}
  \footnotesize
    \begin{tabular}{|c|c|c|c|c|c|}
    \hline
    {\bf Method} & {\bf MAE} & {\bf MSE} & {\bf PARAMS}\\
    \hline
    {Rodriguez et al. \cite{C19}}&655.7&697.8&-\\
    \hline
    {Lempitsky et al. \cite{C10}}&493.4&487.1&-\\
    \hline
    {Idrees et al. \cite{C18}}&419.5&541.6&-\\
    \hline
    {Zhang et al. \cite{C11}}&467.0&498.5&7.1M\\
    \hline
    {CrowdNet \cite{C14}}&452.5&-&14.8M\\
    \hline
    {MCNN \cite{C13}}&377.6&509.1&19.2M\\
    \hline
    {MSCNN}&{\bf 363.7}&{\bf 468.4}&{\bf 2.9M}\\
    \hline
    \end{tabular}
\end{table}

\section{CONCLUSION}
\label{sec:conclusion}
In this paper, we proposed a multi-scale convolutional neural network (MSCNN) for crowd counting. Compared with the recent CNN-based methods, our algorithm can extract scale-relevant features from crowd images using a single column network based on the multi-scale blob (MSB). It is an end-to-end training method with no requirement for multi-column/network pre-training works. Our method can achieve more accurate and robust crowd counting performance with far less number of parameters, which make it more likely to extend to the practical application.

\section{ACKNOWLEDGMENT}
\label{sec:acknowledgement}
This work is supported by the National Natural Science Foundation of China (61171142, 61401163), Science and Technology Planning Project of Guangdong Province of China (2014B010111003, 2014B010111006), and Guangzhou Key Lab of Body Data Science (201605030011).
\bibliographystyle{IEEEbib}
\bibliography{reference}
\end{document}